\documentclass{article} 
\usepackage{nips13submit_e,times}
\usepackage{hyperref}
\usepackage{url}
\usepackage{graphicx}

\usepackage{bm}								
\usepackage{adjustbox}
\usepackage{amsmath}
\usepackage{comment}	
\usepackage{amssymb}
\usepackage{bigints}  
\usepackage{float}
\usepackage[title]{appendix}

\title{A Common Framework for Natural Gradient  and Taylor based Optimisation using Manifold Theory}

\author{
Adnan Haider\\
 Cambridge University Engineering Dept., Trumpington St., Cambridge, CB2 1PZ U.K.\\
Cranberry-Lemon University\\
Pittsburgh, PA 15213 \\
\texttt{hippo@cs.cranberry-lemon.edu} \\
(if needed)\\
}

\usepackage{lineno}

\begin{document}
\nolinenumbers

\maketitle

\begin{abstract}
This  technical report introduces a mathematical framework  to compare and relate  standard Taylor  approximation based optimisation methods with the method of Natural Gradient, an optimisation approach that is Fisher efficient with probabilistic models.  The constructed framework will be shown to also provide  a mathematical justification to combine  higher order Taylor approximation methods with the method of NG.  

\end{abstract}

\section{Introduction}
\vspace{-1em}
The core premise behind first and second order optimisation methods is Taylor's theorem.  Assuming that  the objective function $F(\bm{\theta})$ is sufficiently smooth, Taylor's second order approximation  models the local behaviour of the  function  by the following quadratic function: 

\begin{align}
F(\bm{\theta}_k + \Delta \bm{\theta}) \simeq F(\bm{\theta}_k ) +  \Delta \bm{\theta}^T  \nabla F(\bm{\theta}_k )+ \frac{1}{2} \Delta \bm{\theta}^T H  \Delta \bm{\theta} 
\label{TaylorTheorem}
\end{align}
where $ \Delta \bm{\theta}$ represents any offset  within a convex neighbourhood of   $\bm{\theta}_k$ and $H$ is the Hessian of $F$ computed w.r.t $\bm{\theta}_k$. Instead of optimising the objective function directly, second order methods focus on minimising the above approximate quadratic  at each iteration of the optimisation process. The same approach is undertaken by first order methods but  such approaches make a further  approximation by dropping the quadratic term in Taylor's formula.  This technical report provides a re-derivation of Taylor's theorem from the perspective of manifold theory.   Such a formulation will be shown  to provide a consistent framework to compare the above  approaches with the method of Natural Gradient \cite{Amari1998a,Amari1997,Pascanu2013a,Desjardins2015} for  probabilistic discriminative models $P_{\bm{\theta}}(\mathcal{H}|\mathcal{O})$.  The necessary machinery needed to develop this framework relies on the concept of  manifold, tangent vectors and directional derivatives from the perspective of Information Geometry. Appendix \ref{NB} contains a glossary of terms referenced in this work and   a more in-depth discussion can be found in Amari's text book \cite{AmariBook}.

\section{Deriving Taylor's theorem using manifold theory \label{TaylorTheorem}}
\vspace{-1em}
A   curve on a parameter manifold (Appx. \ref{manifold}) $X$ is a continuous  map  $ c : (a, b)  \subset \mathbb{R} \rightarrow  X$.   Let $\mathcal{U}$ be an open convex neighbourhood  of the current iterate $\bm{\theta}_k$. Thus for any point $\bm{\theta}$ in  $\mathcal{U}$, $\exists$  a curve of the form $\bm{\theta}_k + t (\bm{\theta}-\bm{\theta}_k) $ where $t \in [0,1]$ that is contained in $\mathcal{U}$.

 \vspace{5mm}
Let $ c : [0, 1] \rightarrow  X$  be a  continuous curve such that:
\begin{align}
c(t) = \bm{\theta}_k + t \Delta \bm{\theta} 
\label{curve}
\end{align}
where  $\Delta \bm{\theta}$ corresponds to arbitrary offset from $\bm{\theta}_k$ such that $c$ in contained in $\mathcal{U}$.  The derivative of  $c$ at given point $t_0$ is then the linear map from the tangent space (Appx. \ref{tangent}) at $t_0$ to the tangent space $T_{c(t_0)}  X$:

 \begin{align}
\rm{ d} c  |_{t_0}  \left (   \dfrac{d}{dr } \right) &=  ( \dfrac{\partial}{\partial \hat{\theta}_1}, \dfrac{\partial}{\partial \hat{\theta}_2} \cdots \dfrac{\partial}{\partial \hat{\theta}_D}) \left[
\begin{array}{c}
\Delta \theta_1\\
\Delta \theta_2\\
\vdots\\
\Delta \theta_D
\end{array}
\right] \notag\\
  &=  \sum_i \Delta \theta_i  \dfrac{\partial }{\partial \theta_i}
 \end{align}
 
 where $ \dfrac{d}{dr }$ denotes   the  basis vector  of $T_{t_0}\mathbb{R}$.

 \vspace{2mm}
 Let $F(\bm{\theta})$ be a germ (Appx. \ref{germs}) that corresponds to  a smooth map from the parameter manifold $X$ to $\mathbb{R}$. In the context of optimisation, this corresponds to the smooth objective training criterion.
 The derivative of  $F(\bm{\theta})$ at any given point $\bm{\theta}$ is a  linear map  from the tangent space from  $T_{\bm{\theta}}  X$ to the tangent space  $T_{F(\bm{\theta})}  \mathbb{R}$:
 
 \begin{align}
\rm{ d} F  |_{\bm{\theta}}  ( \sum_i a^{i} \frac{\partial}{\partial \theta_i} |_{\bm{\theta}}) &= \left (   \dfrac{d}{dr } \right)  \left[
( \dfrac{\partial F }{\partial \hat{\theta}_1} |_{\bm{\theta}} , \dfrac{\partial F }{\partial \hat{\theta}_2}|_{\bm{\theta}}  \cdots \dfrac{\partial F}{\partial \hat{\theta}_D}|_{\bm{\theta}} )
\right] 
\left [
\begin{array}{c}
a^1 \\
a^2 \\
\vdots \\
a^D \\
\end{array}
\right ]
 \end{align}
 Here the vector $\nabla F(\bm{\theta}_k )$  is the Jacobian and denotes the particular vector in  $T_{\bm{\theta}} M$ that yields the greatest \textbf{directional} derivative (Appx. \ref{derivative}). 

\vspace{2mm}

Constraining $F$ on the curve $c$ is equivalent to applying  the composite map  $F \circ c$  from $\mathbb{R} \rightarrow \mathbb{R}$. The derivative of such a map   at given point $t_0$ will now correspond to a linear map from a tangent space in $ T_{t_0}\mathbb{R}$ to the tangent space of  $T_{F\circ c(t_0)}\mathbb{R}$.  By applying chain rule such a linear map can be shown to correspond to:
\begin{align}
 \rm{d} (F  \circ  c ) |_{t_0}  \left (   \dfrac{d}{dr } \right)  &= \left (\dfrac{d}{dr} \right) \left [ F  \circ  c \right]_{t_0}   \\
 &= \left (   \dfrac{d}{dr} \right)   \left[
( \dfrac{\partial F }{\partial \hat{\theta}_1} |_{ \bm{\theta}_k + t_0 \Delta \bm{\theta}  } , \dfrac{\partial F }{\partial \hat{\theta}_2}|_{ \bm{\theta}_k + t_0 \Delta \bm{\theta}   }  \cdots \dfrac{\partial F}{\partial \hat{\theta}_D}|_{ \bm{\theta}_k + t_0 \Delta \bm{\theta}   } )
\right] \left[
\begin{array}{c}
\Delta \theta_1\\
\Delta \theta_2\\
\vdots\\
\Delta \theta_D
\end{array}
\right]  \notag\\
&=  \dfrac{d}{dr} \sum_i \Delta \theta_i  \dfrac{\partial F}{\partial \theta_i} |_{ \bm{\theta}_k + t \Delta \bm{\theta}  }
 \end{align}


Using the fact that $F \circ c$ is differentiable and corresponds to a map from  $\mathbb{R} \rightarrow \mathbb{R}$,  by the \emph{ fundamental theorem of calculus}: 

\begin{align}
F(\bm{\theta}_k + \Delta \bm{\theta}) &=  F(\bm{\theta}_k )+  \bigintss_0^1 \left ( F\circ c(t) \right )' dt \notag \\
&=F(\bm{\theta}_k ) + \bigintss_0^{1} \sum_i \Delta \theta_i  \frac{\partial  F}{\partial\theta_{i}} \left (\bm{\theta}_k + t \Delta \bm{\theta} \right) dt  \notag\\
&=F(\bm{\theta}_k ) + \sum_i \Delta \theta_i   \bigintss_0^{1} \frac{\partial  F}{\partial\theta_{i}} \left (\bm{\theta}_k + t \Delta \bm{\theta} \right) dt 
\label{FTC}
\end{align}

Since   individual terms $ \left ( \bigints_0^{1}  \dfrac{\partial  F}{\partial\theta_{i}} ( \bm{\theta}_k + t \Delta \bm{\theta} \mbox {) } dt \right) $  themselves are  smooth functions defined on the convex neighbourhood of $\bm{\theta}_k$, (\ref{FTC}) can be expanded  even further by recursively applying the \emph{ fundamental theorem of calculus}:
\begin{align}
F(\bm{\theta}_k + \Delta \bm{\theta}) &=  F(\bm{\theta}_k ) + \sum_i \frac{\partial F}{\partial \theta_{i}}(\bm{\theta}_k) \Delta \theta_i   + \sum_i \Delta \theta_i  \bigintss_0^{1} \left( \sum_j \Delta \theta_j  \bigintsss_0^{1} \frac{\partial ^2 F}{\partial \theta_{j}\partial \theta_{i}}(\bm{\theta}_k + t \Delta \bm{\theta} ) dt \right ) dt \notag\\
&=F(\bm{\theta}_k ) + \sum_i \frac{\partial F}{\partial \theta_{i}}(\bm{\theta}_k) \Delta \theta_i   + \sum_i \Delta \theta_i   \sum_j \Delta \theta_j  \bigintss_0^{1} \left( \bigintsss_0^{1} \frac{\partial ^2 F}{\partial \theta_{j}\partial \theta_{i}}(\bm{\theta}_k + t \Delta \bm{\theta} ) dt \right ) dt
 \label{FTC2}
\end{align}

As the function $\dfrac{\partial ^2 F}{\partial \theta_{j}\partial \theta_{i}}(\bm{\theta})$ is continuous, when $ \Delta \bm{\theta}$ is sufficiently small,  $\dfrac{\partial ^2 F}{\partial \theta_{j}\partial \theta_{i}}(\bm{\theta}_k + t \Delta \bm{\theta})$ can be approximated by $ \dfrac{\partial ^2 F}{\partial \theta_{j}\partial \theta_{i}}(\bm{\theta}_k + t \Delta \bm{\theta} )\left |_{t =0} \right .$  Under this  approximation, the local behaviour of the germ $F(\bm{\theta})$ can be approximated by:


\begin{align}
F(\bm{\theta}_k + \Delta \bm{\theta}) &\simeq  F(\bm{\theta}_k ) + \sum_i \frac{\partial F}{\partial \theta_{i}} (\bm{\theta}_k) \Delta \theta_i   + \sum_i \Delta \theta_i  \bigintss_0^{1} \left( \sum_j \Delta \theta_j  \bigintsss_0^{1} \frac{\partial ^2 F}{\partial \theta_{j}\partial \theta_{i}}(\bm{\theta}_k ) dt \right ) dt \notag \\
&\simeq  F(\bm{\theta}_k ) + \sum_i \frac{\partial F}{\partial \theta_{i}} (\bm{\theta}_k)\Delta \theta_i   +  \frac{1}{2} \sum_i \Delta \theta_i  \sum_j \Delta \theta_j  \frac{\partial ^2 F}{\partial \theta_{j}\partial \theta_{i}}(\bm{\theta}_k) \label{SecondOrder}
\end{align}

In vector notation this corresponds to:
\begin{align}
F(\bm{\theta}_k + \Delta \bm{\theta}) \simeq  F(\bm{\theta}_k ) +  \Delta \bm{\theta}^T  \nabla F(\bm{\theta}_k )+ \frac{1}{2} \Delta \bm{\theta}^T H  \Delta \bm{\theta} \label{SecondOrder2}
\end{align}
with entries  $H_{j,i}$ of  the Hessian matrix corresponding to $\left(  \dfrac{\partial ^2 F}{\partial \theta_{j}\partial \theta_{i}}(\bm{\theta}_k)  \right)$.   The above expression corresponds to Taylor's second order approximation. Re-deriving this expression from the perspective of manifold theory shows how the  product $\Delta \bm{\theta}^T  \nabla F(\bm{\theta}_k )$ can be interpreted as an \textbf{inner product} between vectors $ \Delta \bm{\theta} $ and $ \nabla F(\bm{\theta}_k) $ in $T_{\bm{\theta}} X$. This interpretation will become handy in the next section. 

\section{ Formulating Taylor's quadratic as a  minimisation problem in the Tangent space \label{TS}}
\vspace{-1em}
Instead of   minimising the objective function directly, second order methods   focus on minimising the  quadratic function of (\ref{SecondOrder2}) at each iteration. The quadratic corresponds to a local model of the behaviour of $F(\bm{\theta})$ within a convex neighbourhood of the current iterate $\bm{\theta}_k$. By re-deriving Taylor's second order approximation from the perspective of manifold theory in the previous section, it was shown how $\Delta \bm{\theta} $ corresponds to a particular choice of  a tangent vector from $T_{\bm{\theta}_k} X$, and  $\nabla F(\bm{\theta}_k )$ represents the particular vector in  $T_{\bm{\theta}} X$ that yields the greatest directional derivative under the linear map $\rm{d} F  |_{\bm{\theta}}$ . Thus,  as  $ F(\bm{\theta}_k )$ is a  constant,  solving the minimisation problem in (\ref{SecondOrder2}) is equivalent to solving the following minimisation problem in  $T_{\bm{\theta}_k} X$:

\begin{align}
\Delta \hat{\bm{\theta}} =  \arg\min_{\Delta \bm{\theta} \in T_{\bm{\theta}_k} X}  F(\bm{\theta}_k ) +  \langle \Delta \bm{\theta},  \nabla F(\bm{\theta}_k ) \rangle + \frac{1}{2} \Delta \bm{\theta}^T H  \Delta \bm{\theta} 
\label{IHF}
\end{align}
where $  \langle \Delta \bm{\theta},  \nabla F(\bm{\theta}_k ) \rangle$  corresponds to the standard inner product between vectors in $T_{\bm{\theta}_k} X$   and  $ \Delta \bm{\theta}^T H  \Delta \bm{\theta}$ corresponds  to a linear map $g : \bm{u} \in  T_{\bm{\theta}_k} X \rightarrow \mathbb{R}$. 

\subsection{Relating Gradient Descent with Natural Gradient \label{SGDNG}}
\vspace{-1em}
Under this framework, first order methods can be seen to solve the following optimisation problem in the  
tangent space $T_{\bm{\theta}_k} X$:
\begin{align}
\Delta \hat{\bm{\theta}} =  \arg\min_{\Delta \bm{\theta} \in T_{\bm{\theta}_k} X}  F(\bm{\theta}_k ) +  \langle \Delta \bm{\theta},  \nabla F(\bm{\theta}_k ) \rangle 
\label{FIT}
\end{align}

 Since $X$ is a manifold,  the inner product endowed on the tangent space  $T_{\bm{\theta}} X$ at any point $\bm{\theta}$ need not be just the identity matrix. The parameter manifold $X$ can be equipped with any form of a  Riemannian metric,  a smooth map that assigns to each $\bm{\theta} \in X$ an inner product  $I_{\bm{\theta}}$  in $T_{\bm{\theta}} X$.  When the underlying model  corresponds to a discriminative probabilistic model $P_{\bm{\theta}}(\mathcal{H}|\mathcal{O})$, a particular choice  of  $I_{\bm{\theta}}$ is the Fisher Information matrix \cite{Amari1998a,Amari1997}:

\begin{align}
I_{\bm{\theta}}  &=  E_{ P_{\bm{\theta}}(\mathcal{H}|\mathcal{O}) }  \left [ \left( \nabla \mbox { log } P_{\bm{\theta}}(\mathcal{H}|\mathcal{O}) \right ) \left( \nabla \mbox { log } P_{\bm{\theta}}(\mathcal{H}|\mathcal{O}) \right )^T \right ] 
 \label{equf}    
\end{align}
Equipping the tangent space $T_{\bm{\theta}} X$ with the above Riemannian metric  allows interpretation of  lengths of paths traversed in the parameter space as changes in the KL divergence (Appx. \ref{KLFI} ). When $X$  possesses such a structure, performing  first order optimisation within a trust region defined by $I_{\bm{\theta}}$ can be shown to produce the  update $\Delta \bm{\theta} =I_{\bm{\theta}}^{-1}  \nabla F(\bm{\theta} )$ at each iteration \cite{Bottou2016,Adnan2017}.  From the perspective of  Information Geometry, this corresponds to the direction of steepest descent in the space of $P_{\bm{\theta}}(\mathcal{H}|\mathcal{O})$.  Thus, in this sense,  such an update is termed as the direction of Natural  Gradient (NG) \cite{Amari1998a,Amari1997,Pascanu2013a,Desjardins2015} .

Using the fact that  $I_{\bm{\theta}}$  is symmetric and positive definite, it is also possible  to endow $X$ with a Riemannian metric of the form   $I_{\bm{\theta}}^{-1}$  by the \emph{spectral decomposition theorem}. With respect to such a metric,  solving the minimising problem of (\ref{FIT}) in the tangent space $ T_{\bm{\theta}_k} X$  becomes equivalent to performing NG on the parameter surface. Hence, recasting the optimisation problem to a minimisation problem  in $ T_{\bm{\theta}_k} X$ provides a nice framework to relate  Gradient Decent with the method of NG. 

In practice, since it is not feasible to compute the expected  outer product of the likelihood score (\ref{equf}) exactly,  $I_{\bm{\theta}}$  is approximated  by its Monte-Carlo estimate $\hat{I}_{\bm{\theta}}$.  As such a matrix  is only positive semi-definite, it's inverse is not guaranteed to exist. To address this issue,  this work also provides the derivation of  an  alternative dampened Riemannian metric $ \tilde{I}_{\bm{\theta}}^{-1}$ (Appx. \ref{Proof}) that  is not only guaranteed to be positive definite but has the feature that its image space is the direct sum of the image  and  the kernel space of the empirical Fisher matrix $\hat{I}_{\bm{\theta}}^{-1}$. Assigning such a metric has one particular advantage: the very first directions explored by the CG algorithm constitute to directions in the image space of $\hat{I}_{\bm{\theta}}$ \cite{NocedalBook} i.e  directions considered important by the empirical Fisher are  traversed first during the initial stages of a CG run.

\section{Summary}
To summarise, this report  presents a  mathematical framework  to compare and relate  standard Taylor  approximation based optimisation methods with the method of Natural Gradient, an optimisation approach that is Fisher efficient with probabilistic models. The constructed framework can be seen to also provide  a mathematical justification to combine  higher order Taylor approximation methods with the method of NG.



\begin{appendices}
\section{Glossary \label{NB}}
\vspace{-1em}
This section begins with  a formal description of the concept of a smooth manifold. Conceptually, a manifold corresponds to any geometric object embedded in $\mathbb{R}^k$ that is locally Euclidean i.e any local patch of the object is  topologically equivalent to an open unit ball in a smaller dimensional  Euclidean space. In $\mathbb{R}^3$, curves and surfaces are  examples of  an embedded manifold. When movement is  constrained  only along  these objects, the degrees of freedom with which one can  traverse  is lower than dimensionality of the embedded 3 dimensional space.

\subsection{ Formal definition of  a manifold \label{manifold}}
A topological manifold $X$ of dimension $D$ is a second countable Hausdorff topological space that is locally homeomorphic to $\mathbb{R}^D$;  that is, for any point $\bm{\theta} \in X$, there exists an open neighbourhood $U$ of  $\bm{\theta}$ and a homeomorphism $g : U \rightarrow O \subset \mathbb{R}^D$, where $O$ is open in $\mathbb{R}^D$. We call the homeomorphism $g : U \rightarrow O$ a chart, and the neighbourhood $U$ a coordinate neighbourhood of $\bm{\theta}$.  

\vspace{5mm}
To clarify the reader, in topology the concept  of  a homeomorphism  is equivalent  to a bijective continuous map and  the concept of Hausdorff means for any two points $\hat{\bm{\theta}},\bar{\bm{\theta}} \in X $, it is always possible to find disjoint open neighbourhoods that contain each point.

\vspace{1mm}
  Let $ r_i: \mathbb{R}^D \rightarrow \mathbb{R}$ denote the projection onto the ith coordinate. Given a chart $g : U \rightarrow O \subset \mathbb{R}^D$, let $ \theta_i : r_i \circ g : U \rightarrow \mathbb{R}$. The functions $\theta_i$ are  the local coordinates w.r.t the chart $g$ on $U$.  Having established a notion of what it means to be manifold, 
the next concept that will be introduced is the concept of a tangent vector. To formally define the concept of a tangent vector at any point $\bm{\theta} \in X$, it is first necessary to formalise the notion of \emph{germs} associated with a given point in  a manifold.

\subsection{Germs \label{germs}}
Let $X$ be a manifold and $\bm {\theta} \in X$. Functions $f$, $g$ defined on open subsets $U$, $V$ respectively containing $\bm{\theta}$ are said to have the same germ at $\bm{\theta}$  if there exists a neighbourhood $W$ of $\bm{\theta}$ contained in $ U \cap V$ such that $f |_{W} \equiv g |_{W}$.  The notion of germs 
therefore defines an equivalence relation on the space of functions defined on an open neighbourhood of  $\bm{\theta}$ where  $ (U, f) \sim (V, g)$ if and only if there exists
a neighbourhood $W$ of $\bm{\theta}$ contained in $ U \cap V$ such that $f |_{W} \equiv g |_{W}$.  Let $C^{\infty}_{\bm{\theta}}$ be the set of all such equivalent function classes.  Having defined the concept of class of germs associated  with a given point $\bm {\theta} \in X$, the necessary machinery is now in place to define the concept of a tangent vector associated with the point $\bm {\theta} \in X$.

\subsection{Tangent vector\label{tangent}}
A tangent vector $\bm{v}$ at  given point $\bm{\theta} \in X$  is a linear derivation of  $C^{\infty}_{\bm{\theta}}$, that is it a special form of  a linear map $C^{\infty}_{\bm{\theta}} \rightarrow \mathbb{R}$  that satisfies the property $\bm{v}(f \cdotp g) = f(\bm{\theta})\bm{v}(g) + \bm{v}(f)g(\bm{\theta})$ where $f \cdotp g$ denotes the product of functions $f \mbox{ and } g  \in C^{\infty}_{\bm{\theta}}$. The tangent vectors form a real vector space in the obvious way;  this space is denoted by $T_{\bm{\theta}}(X)$ and is called the tangent space to $X$ at $\bm{\theta}$. 

\vspace{5mm}

The concept of tangent vectors is necessary  to do calculus on manifolds. Since manifolds are locally Euclidean, the usual notions of differentiation and integration make sense in any coordinate chart and  can be carried over to manifold space. Specifically, a tangent vector as will be shown now is the manifold version of a directional derivative at a point $\bm {\theta} \in X$.

\textbf{ Basis for Tangent space \label{basis}}
Let $(\theta_i,\theta_2,\cdots \theta_D)$    denote the standard coordinates of the parameter space $X$. Consider the operator $\frac{\partial}{\partial \theta_i} |_{\bm{\theta}}$ defined   by $ \dfrac{\partial}{\partial \theta_i }|_{\bm{\theta}} (f) = \dfrac{\partial f}{\partial \theta_i}({\bm{\theta}})$. Then the  set $\left \{\frac{\partial}{\partial \theta_i} |_{\bm{\theta}}  \right \}_{i}$ can be shown to be  linear derivations of  $C^{\infty}_{\bm{\theta}}$ and hence members of  $T_{\bm{\theta}}(X)$. Furthermore,  for $\forall \theta_j$, each  operator in this set satisfies  
\[ \frac{\partial}{\partial \theta_i} |_{\bm{\theta}}({\theta_j})  =
 \begin{cases}
 1  &  i \equiv j   \\
 0   & \text{ otherwise}\\ 
\end{cases}\]
It can be shown that, by satisfying the above constraint,   the members of $\{\frac{\partial}{\partial \theta_i} |_{\bm{\theta}}\}_i$ correspond to a basis of  $T_{\bm{\theta}}(X)$.  Therefore, w.r.t this basis  if  $\bm{v} \in T_{\bm{\theta}}(X) $ then  $\bm{v} = \sum_i a^{i} \frac{\partial}{\partial \theta_i} |_{\bm{\theta}}$.

\subsection{Concept of directional derivative\label{derivative}}
Let $\Phi: X \rightarrow N $  be a  vector valued function that corresponds to a smooth map between two manifolds. The directional derivative of $\Phi$ at  any point  $\bm{\theta} \in X$ is  the linear map 
\begin{align*}
d\Phi |_{\bm{\theta}} : T_{\bm{\theta}}(X) \rightarrow T_{\Phi(\bm{\theta})}(N)
\end{align*}
defined by:
\begin{align}
\rm{d}\Phi |_{\bm{\theta}}( \sum_i a^{i} \frac{\partial}{\partial \theta_i} |_{\bm{\theta}}) = \left (\frac{\partial}{\partial y_i} |_{ \Phi (\bm{\theta})},\frac{\partial}{\partial y_i} |_{ \Phi (\bm{\theta})},\cdots \frac{\partial}{\partial y_k} |_{ \Phi (\bm{\theta})} \right ) \left[
\begin{array}{ccc}
 \frac{\partial \Phi_1}{\partial \theta_1} |_{\bm{\theta}} &  \frac{\partial \Phi_1}{\partial \theta_2} |_{\bm{\theta}_2} \cdots   \\ \\
\frac{\partial \Phi_2}{\partial \theta_1} |_{\bm{\theta}} &  \frac{\partial \Phi_2}{\partial \theta_2} |_{\bm{\theta}_2} \cdots \\
\vdots\\
\frac{\partial \Phi_k}{\partial \theta_1} |_{\bm{\theta}} &  \frac{\partial \Phi_k}{\partial \theta_2} |_{\bm{\theta}_2} \cdots
\end{array}
\right] 
\left [
\begin{array}{c}
a^1 \\
a^2 \\
\vdots \\
a^D \\
\end{array}
\right ]
\end{align}
where   $\{\frac{\partial}{\partial y_j} |_{\Phi(\bm{\theta})}\}_j$  denotes the basis of $ T_{\Phi(\bm{\theta})}(N)$. Each coordinate of $\frac{\partial}{\partial y_j} |_{\Phi(\bm{\theta})}$ corresponds to the directional derivative of $\Phi_j$ w.r.t $\bm{\theta}$. 

\vspace{2mm}

\section{}
\subsection{Approximating the KL divergence using the Fisher Information Matrix \label{KLFI}}
\vspace{-1em}
Let ${X}$  denote  the  parameter manifold.  As different realisations of model parameters lead to different probabilistic models  $P_{\bm{\theta}}(\mathcal{H}|\mathcal{O})$, the manifold  is homeomorphic or in other words equivalent to the space of all probability distributions $\mathcal{M}$ that can be captured by the chosen model. The goal of learning is to identify  a viable candidate ${h}(\bm{\theta}|\mathcal{O}) \in \mathcal{M}$ that  avoids rote memorisation and instead generalises to  the concepts that can be learned from a given set of examples. 

Using the fact that each candidate  ${h}(\bm{\theta}|\mathcal{O})$  is a valid probability distribution, the derivatives of such densities satisfy the identity: 
\begin{align}
   \sum_H  \frac{\partial}{\partial \theta^i} {h}(\bm{\theta}|\mathcal{O})  =  \sum_H  \frac{\partial}{\partial \theta_i}P_{\bm{\theta}}(\mathcal{H}|\mathcal{O})  =\frac{\partial}{\partial \theta_i}   \sum_H P_{\bm{\theta}}(\mathcal{H}|\mathcal{O})  =   \frac{\partial}{\partial \theta_i}  1 =0 
 \label{Axiom}  
\end{align}

Since  ${X}$ is homeomorphic to $\mathcal{M}$, adding a small quantity $\Delta \bm{\theta}$ to the current iterate $ \bm{\theta}$  results in a unique distribution 
$P_{\bm{\theta}+ \Delta \bm{\theta}}(\mathcal{H}|\mathcal{O})$. To quantify the degree of separation between this point from the previous point  $P_{\bm{\theta}}(\mathcal{H}|\mathcal{O})$ in $\mathcal{M}$,  a standard  divergence measure for probabilistic models is  the KL-divergence $KL \left (P_{\bm{\theta}}(\mathcal{H},|\mathcal{O}) \|P_{\bm{\theta}+\Delta \bm{\theta}}(\mathcal{H}|\mathcal{O})\right)$.

 The KL-divergence is  a functional that maps the space of  distributions  $\mathcal{M}$ to $\mathbb{R}$.  Since each distribution itself is a function of $\bm{\theta}$, the divergence measure can be interpreted as a smooth function of $\bm{\theta}$. This allows the local  behaviour of this divergence measure to be approximated within a convex neighbourhood of the current point $\bm{\theta}$ by Taylor's second order approximation:
 
\begin{align}
KL \left (P_{\bm{\theta}}(\mathcal{H}|\mathcal{O}) \|P_{\bm{\theta}+\Delta \bm{\theta}}(\mathcal{H}|\mathcal{O})\right)  =&E_{P_{\bm{\theta}}(\mathcal{H}|\mathcal{O})} \left [  \mbox{ log } P_{\bm{\theta}}(\mathcal{H}|\mathcal{O})  - \mbox{ log } P_{\bm{\theta}+\Delta \bm{\theta}}(\mathcal{H}|\mathcal{O})  \right ]  \notag \\
\approx &   - \frac{1}{2} \Delta \bm{\theta}^T  E_{P_{\bm{\theta}}(\mathcal{H}|\mathcal{O})} \left [  \nabla^2  \mbox { log }  P_{\bm{\theta}}(\mathcal{H}|\mathcal{O})\right ]\Delta \bm{\theta}  
\label{KLTaylor}
\end{align}

In (\ref{KLTaylor}), the first order term is dropped as it equates to zero by (\ref{Axiom}),  Hence, locally the KL divergence can be approximated  by the bilinear form:

\begin{align}
KL \left (P_{\bm{\theta}}(\mathcal{H}| \mathcal{O}) \|P_{\bm{\theta}+\Delta \bm{\theta}}(\mathcal{H}| \mathcal{O})\right)  \approx &- \frac{1}{2} \Delta \bm{\theta}^T  E_{P_{\bm{\theta}}(\mathcal{H}| \mathcal{O})} \left [  \nabla^2  \mbox { log }  P_{\bm{\theta}}(\mathcal{H}|\mathcal{O})\right ]\Delta \bm{\theta} 
\label{KLdiv}
\end{align}
For discriminative models $P_{\bm{\theta}}(\mathcal{H}|\mathcal{O})$,  the Fisher Information  for a random variable $\bm{\theta}$  corresponds to  the expected  outer product of  \textbf{score} of the likelihood:

\begin{align*}
I_{\bm{\theta}} &=    E_{P_{\bm{\theta}}(\mathcal{H}|\mathcal{O})} \left [ \left( \nabla \mbox { log }  {P}_{\bm{\theta}}(\mathcal{H}|\mathcal{O}) \right ) \left( \nabla \mbox { log }  {P}_{\bm{\theta}} (\mathcal{H}|\mathcal{O}) \right )^T \right ]  \\
\end{align*}

 In scenarios where  (\ref{Axiom}) holds, $I_{\bm{\theta}}$  can be shown to be equal to the negative of the expectation of the Hessian w.r.t the distribution $P_{\bm{\theta}}(\mathcal{H}|\mathcal{O})$. Thus by substituting the expression of $I_{\bm{\theta}}$ into (\ref{KLdiv}), the KL divergence measure is locally equivalent to the inner product:
\begin{align}
KL \left (P_{\bm{\theta}}(\mathcal{H}|\mathcal{O}) \|P_{\bm{\theta}+\Delta \bm{\theta}}(\mathcal{H}|\mathcal{O})\right) \approx \frac{1}{2} \Delta \bm{\theta}^T I_{\bm{\theta}}  \Delta \bm{\theta}
\end{align}

Let  $ \alpha : (a, b)  \subset \mathbb{R} \rightarrow  X$ be a continuous map on $X$.  Assuming that the curve is regular i.e $\rm{ d} \alpha$ is injective at every point $t \in(a,b)$, the arc length of the curve corresponds to:
\begin{align*} s(t) &= \bigintss_a^b \| \left \langle \rm{ d} \alpha(t) ,\rm{ d} \alpha(t) \right \rangle_{I_{\bm{\theta}}} \| dt  
\end{align*}
If the choice of  $I_{\bm{\theta}}$ corresponds to the Fisher Information matrix,  length of paths traversed in the parameter manifold can be interpreted as  local changes in the KL divergence.

\section{ Adapting the empirical Fisher to  a yield  a proper Riemannian metric  \label{Proof}}

To recap, a Riemannian metric on a smooth parameter manifold $X$ (see Appendix \ref{manifold}) is  a smooth map that assigns to each $\bm{\theta} \in X$ an inner product  $I_{\bm{\theta}}$  in $T_{\bm{\theta}} X$. As   $\hat{I}_{\bm{\theta}}$  is real and symmetric, by the  \emph{spectral decomposition theorem} \cite{Friedberg}, there exists a unitary basis of eigenvectors w.r.t the matrix becomes diagonalisable:
\begin{align}
 \hat{I}_{\bm{\theta}} \equiv V_{\bm{\theta}} \Lambda_{\bm{\theta}} V^T_{\bm{\theta}} 
\end{align} 
where   $ V_{\bm{\theta}} $ is a square matrix whose $i$-th column  corresponds to the $i$-th eigenvector of $\hat{I}_{\bm{\theta}}$,  and $ \Lambda_{\bm{\theta}} $ is the diagonal matrix whose non-zero entries represent the associated eigenvalues. It should be noted that the entries of these two matrices  are functions of $\bm{\theta}$. To  keep the  notation uncluttered, the dependency on $\bm{\theta}$ will be dropped for the remainder of this section whenever any of the individual factors in $V_{\bm{\theta}} \Lambda_{\bm{\theta}} V^T_{\bm{\theta}}$ is mentioned. Since $\hat{I}_{\bm{\theta}}$  is only guaranteed to be positive semi-definite, there will  exist  zero diagonal entries in $\Lambda$ resulting in its rank being $m$ where $m< D$ (the dimensionality of the parameter space).  Under such circumstances, $ \hat{I}_{\bm{\theta}}^{-1}$ will not exist and  it will no longer be possible to endow $X$ with Riemannian metric of the form $ \hat{I}_{\bm{\theta}}^{-1}$. To address this issue, this section derives an alternative metric to $ \hat{I}_{\bm{\theta}}$ that has the same structure and properties as the damped FI matrix but is guaranteed to be positive definite. From a high level perspective, the construction of the proposed  metric is achieved in two stages.

\textbf{Step} 1: partition the tangent space $T_{\bm{\theta}}  X$ into  two disjoint subspaces such that one  subspace is spanned by eigenvectors of $ \hat{I}_{\bm{\theta}}$ associated with  non-zero eigenvalues.\\ 
 By re-arranging the columns of $V$ such that the eigenvectors associated with the non-zero eigenvalues occupy positions within the first $m$  columns,  the image space of   $\hat{I}_{\bm{\theta}}$ can be then essentially  captured by the matrix $V_{:,1:m} \Lambda_{1:m,1:m} V^T_{:,1:m}$. 

Let  $\phi$ be a map from $X$ to $\mathbb{R}^m$ given by 
\begin{align}
\phi (\bm{\theta}) =  V^T_{:,1:m} \bm{\theta}
\end{align}

By  the \emph{Replacement theorem} \cite{Friedberg}, such a map can be shown to be maximal i.e the derivative $\rm{d} \phi$ is of full rank.  Under the \emph{Implicit Function theorem} \cite{kline1998calculus}, there exists a chart $h$ (see Appendix \ref{manifold})  on  $X$ and a neighbourhood  $\mathcal{V}$ of  $h(\bm{\theta})$ such that  $\phi \circ  h^{-1} |_{\mathcal{V}} = \pi|_{\mathcal{V}}$ with  $ \pi : \mathbb{R}^D \rightarrow \mathbb{R}^m$ being the projection map. It is easy to see that  from the  definition of $\phi$ that such a chart does exist and corresponds to $V_{:,1:m}$. Thus, within the open neighbourhood  $\mathcal{V}$, the derivative of $\phi \circ  h^{-1}$  corresponds to the linear map:
\begin{align}
 \rm{d} \left( \phi \circ  h^{-1} \right)(\bm{v})= \left [ \begin{array}{c|c}
\text{\textbf{I}}_{m \times m} & 0_{(m+1) \times D}\\
\end{array}
\right ] ( \bm{v}) 
\end{align}
With respect to the map $\phi$, the tangent space $T_{\bm{\theta}} X$ can thus be expressed the disjoint sum: 
\begin{align} T_{\bm{\theta}_k} X = T_{ \phi \circ  h^{-1}(\bm{\theta})} \mathbb{R}^m \bigoplus \rm{ker} \left(\rm{d} ( \phi \circ  h^{-1}) \right) \end{align}
where  $\rm{ker} \left(\rm{d} ( \phi \circ  h^{-1}) \right)$ denotes the null space\footnote{ the $\rm{ker}$ of a linear map $L : N \rightarrow W$ between two vector spaces $N$ and $W$, is the set of all elements $\bm{v} \in N$ for which $ L(\bm{v}) = 0$}. By the above construction,  $T_{ \phi \circ  h^{-1}(\bm{\theta})} \mathbb{R}^m$ can  now be identified  as a subspace of $T_{\bm{\theta}} X$.

 \textbf{Step} 2: assign the identity matrix scaled  by a very small number $\epsilon$ to the  tangent subspace captured by the  kernel of $\rm{d} ( \phi \circ  h^{-1}) $. As the entries in the  diagonal  of $ \Lambda_{1:m,1:m}$  are  functions of $\bm{\theta}$, endowing  $T_{ \phi \circ  h^{-1} (\bm{\theta})} \mathbb{R}^m$ with the inner product  $\Lambda_{1:m,1:m}$ is equivalent to assigning a Riemannian metric to the associated subspace in $T_{\bm{\theta}} X$. Together,  the tangent space of $X$  can now be assigned   with the  following Riemannian metric:
 \[ \left[
\begin{array}{c|c}
\Lambda_{1:m,1:,m} & 0\\
0 &\epsilon \text{\textbf{ I}}_{(D-m) \times (D-m)}\\
\end{array}
\right]
\]
Switching back to the original basis coordinates, the proposed  metric then takes the particular form:
\begin {align}
 V \left[
\begin{array}{c|c}
\Lambda_{1:m,1:m} & 0\\
0 & \epsilon \text{\textbf{ I}}_{(D-m) \times (D-m)}\\
\end{array}
\right] V^T  = \tilde{I}_{\bm{\theta}}
\end{align}

It can be seen that by construction  $\tilde{I}_{\bm{\theta}}$  corresponds to a positive definite matrix whose  image space is the direct sum of the image  and  the kernel space of the empirical Fisher matrix $\hat{I}_{\bm{\theta}}$. Apart from being a proper Riemannian metric, using CG to solve the appropriate linear system  $\tilde{I}_{\bm{\theta}} \Delta \bm{\theta} = -\nabla F(\bm{\theta}_k )$, has one particular advantage: the very first directions explored by the CG algorithm constitute to directions in the image space of $\hat{I}_{\bm{\theta}}$ \cite{NocedalBook} i.e  directions considered important by the empirical Fisher are  traversed first during the initial stages of a CG run.

\end{appendices}


\bibliographystyle{IEEEtran}

\end{document}